\documentclass[10pt,twocolumn,letterpaper]{article}

\usepackage[pagenumbers]{iccv} 

\usepackage[ruled, vlined]{algorithm2e}
\usepackage{algpseudocode}
\usepackage{booktabs}
\usepackage{multirow}
\usepackage{float}
\usepackage{setspace}
\DeclareMathOperator*{\argmin}{argmin}

\definecolor{iccvblue}{rgb}{0.21,0.49,0.74}
\usepackage[pagebackref,breaklinks,colorlinks,allcolors=iccvblue]{hyperref}

\def\paperID{138} 
\def\confName{ICCV}
\def\confYear{2025}

\title{Humans as a Calibration: Dynamic 3D Scene Reconstruction from Unsynchronized and Uncalibrated Videos}

\author{Changwoon Choi$^{1*}$\and Jeongjun Kim$^1$\and Geonho Cha$^2$\and Minkwan Kim$^1$\and Dongyoon Wee$^2$\and Young Min Kim$^{1\dagger}$\\
{\small$^1$Seoul National University$\quad$ $^2$NAVER Cloud}
}

\begin{document}

\twocolumn[{
\renewcommand\twocolumn[1][]{#1}
\maketitle
\begin{center}
    \centering
        \captionsetup{type=figure}
        \includegraphics[trim={0 5mm 0 2mm}, clip,width=\linewidth]{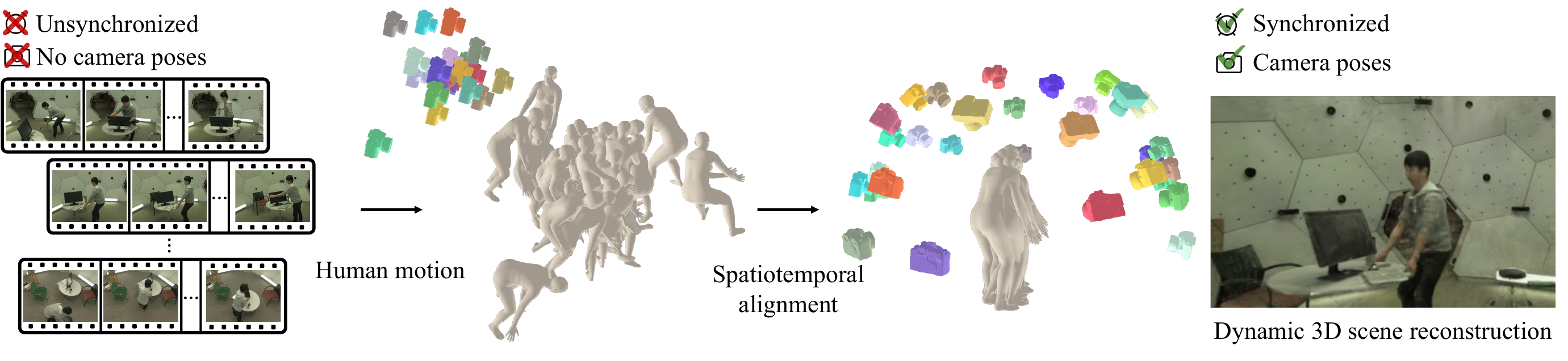}
        \captionof{figure}{We propose an approach to reconstruct dynamic 3D scenes from unsynchronized and uncalibrated videos. We exploit human motion as a calibration pattern to align time offsets and camera poses.} 
        \label{fig:teaser}
\end{center}
}]
\maketitle
\def\thefootnote{*}\footnotetext{Work started during Changwoon's internship at NAVER.}
\def\thefootnote{$\dagger$}\footnotetext{Young Min Kim is the corresponding author.}
\def\thefootnote{\arabic{footnote}}

\begin{abstract}
Recent works on dynamic 3D neural field reconstruction assume the input from synchronized multi-view videos whose poses are known.
The input constraints are often not satisfied in real-world setups, making the approach impractical.
We show that unsynchronized videos from unknown poses can generate dynamic neural fields as long as the videos capture human motion.
Humans are one of the most common dynamic subjects captured in videos, and their shapes and poses can be estimated using state-of-the-art libraries.
While noisy, the estimated human shape and pose parameters provide a decent initialization point to start the highly non-convex and under-constrained problem of training a consistent dynamic neural representation.
Given the shape and pose parameters of humans in individual frames, we formulate methods to calculate the time offsets between videos, followed by camera pose estimations that analyze the 3D joint positions.
Then, we train the dynamic neural fields employing multiresolution grids while we concurrently refine both time offsets and camera poses.
The setup still involves optimizing many parameters; therefore, we introduce a robust progressive learning strategy to stabilize the process.
Experiments show that our approach achieves accurate spatio-temporal calibration and high-quality scene reconstruction in challenging conditions. \href{http://changwoonchoi.github.io/HCP}{Project page}
\end{abstract}    

\section{Introduction}
\label{sec:intro}

\begin{figure}
    \centering
    \includegraphics[trim={0, 5mm, 0, 0},clip,width=0.97\linewidth]{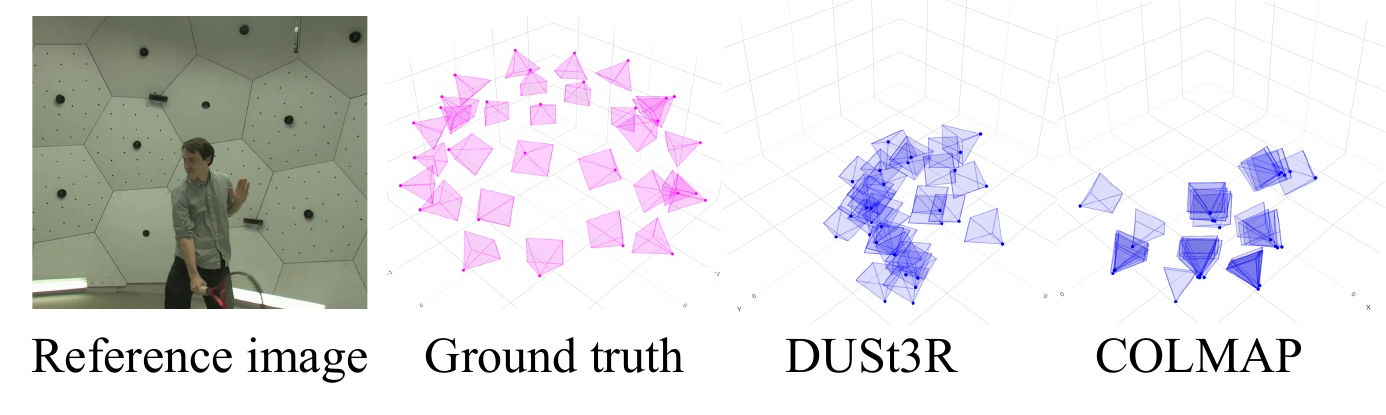}
    \caption{Both transformer-based method DUSt3R~\cite{dust3r} and SfM library COLMAP~\cite{colmap} fail to recover correct camera poses.}
    \vspace{-1em}
    \label{fig:pose_failure}
\end{figure}

Recent advances in neural radiance fields (NeRF) have been extended to dynamic scenes.
However, even with a series of regularization techniques, obtaining a dynamic 3D volume with video inputs is a highly under-constrained problem and often suffers from instability and convergence problems.
The claimed success of dynamic NeRF assumes unrealistic movements of a monocular camera that can emulate a multi-view setup~\cite{dycheck} or perfectly calibrated and synchronized multi-view inputs.
Such input constraints are hard to achieve in the real world.
Time synchronization often relies on hard-wired devices~\cite{panoptic,human36m} or needs additional cues~\cite{sloper4d} (e.g., sound).
Pose estimation methods struggle in scenes with textureless areas or repetitive structures, as shown in~\cref{fig:pose_failure}.
Furthermore, it becomes more challenging if scenes are dynamic and videos are not synchronized.

We aim to allow training dynamic NeRF from a set of casually captured real-world videos of a shared event, such as a sports game or a concert, which can only be collected later.
In other words, we reconstruct photorealistic 4D scenes from \emph{unsynchronized} multi-view videos with \emph{unknown camera poses}.
We need strong priors to tackle this challenging problem with high degrees of freedom.
We focus on \emph{humans}, one of the most common dynamic objects in scenes.
Human pose estimation in computer vision has progressed rapidly and now achieves reliable performance even in general images or videos.
We consider an estimated human parameter a robust mid-level representation to deduce the relationship between unsynchronized videos.
Namely, we exploit human motion as a calibration pattern to estimate both time offsets and camera poses. 

We first find the initial time offsets and camera poses using the sequence of human shape and pose parameters estimated from individual videos. 
We can obtain a robust estimation of time offsets by enforcing global consistency in the sequence of motions with pairwise scores.
Specifically, we consider the estimated parametric model of humans as a time series whose feature vector is the pose and shape parameters.
We apply dynamic time warping (DTW) between every pair of videos to compute a constant offset to minimize the distance between 3D joint positions and record the associated cost value.
The pairwise values are stored within a matrix, from which we find a global sequence alignment with the overall minimum costs.
We estimate camera poses at coherent world coordinates from the alignment of human motions, by applying Procrustes analysis on aggregated 3D joint positions across overlapping frames.

Even with slight misalignments of camera parameters, NeRF reconstruction degrades significantly.
Therefore, we further refine the estimated time offsets and camera poses by jointly optimizing them with dynamic NeRF~\cite{kplanes} training.
We modify coarse-to-fine registration~\cite{barf} to stabilize the training of the multiresolution grid-based representation.
Furthermore, we propose a curriculum learning strategy that adds more variables as the optimization progresses.
Optimization scheduling is critical for convergence in gradient-based optimization.

We evaluate our approach in various challenging real-world datasets including CMU Panoptic Studio~\cite{panoptic}, Mobile-Stage~\cite{4k4d}, and EgoBody~\cite{egobody}.
Our initialization stage with human motion alignment robustly estimates both time offsets and camera poses.
Also, refinement with joint optimization achieves both highly accurate calibration and high-quality dynamic 3D scene reconstruction.

\section{Related Works}
\label{sec:related_works}
\paragraph{Human pose and shape estimation}
Recent human pose and shape estimation methods leverage powerful data-driven prior from abundant datasets containing humans~\cite{lsp,amass,mpii,mscoco,human36m}.
Most approaches recover human pose and shape by estimating parametric models such as SMPL~\cite{smpl}.
From optimization-based methods~\cite{smplify,smplx} to regression-based methods~\cite{hmr,hmmr,vibe}, early works faithfully recover 3D human pose and shape in camera coordinate.
Recent advances~\cite{pace,trace} take a step forward to estimate the global trajectories of humans, which is crucial for a complete understanding of human motion.
Notably, SLAHMR~\cite{slahmr} utilizes human motion priors to recover both real-world scaled global human motion and camera trajectory, whose scale is ambiguous with the SLAM pipeline alone.
We note that monocular video human shape and motion estimation works have matured and generalize well to novel scenes.
In this paper, we explore the potential of human motion as a \emph{robust mid-level representation} for multiple camera calibration in both temporal and spatial domain.

\begin{figure*}
    \centering
    \includegraphics[trim={0, 8mm, 5mm, 0}, clip, width=\linewidth]{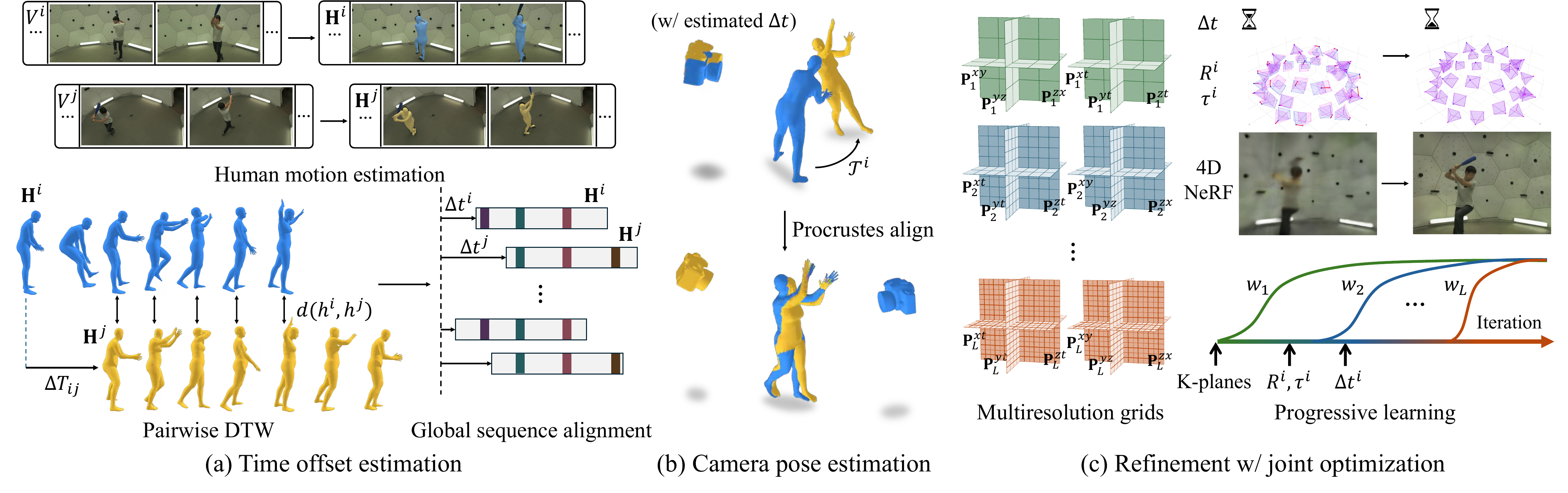}
    \caption{Overview of our method.
    Given unsynchronized multi-view videos without camera poses, we first extract human motion independently.
    Then we estimate (a) time offsets and (b) global camera poses by aligning human motions.
    Starting from the initial point, (c) we further refine both camera poses and time offsets by jointly optimizing them with dynamic NeRF with progressive training.
    }
    \label{fig:method_overview}
\end{figure*}

\paragraph{Camera calibration with human}
While the majority of camera calibration methods exploit image features, there are several recent works that utilize humans as a calibration cue.
For example, Patel and Black~\cite{patel2024camerahmr} propose a single image camera intrinsics estimation network trained on a dataset of images containing people.
Ma et al.~\cite{ma2022virtual} propose a bundle adjustment method with Virtual Correspondences between image pair.
Xu et al.~\cite{xu2021wide} utilize center of bounding boxes of detected humans as keypoints for wide-baseline camera calibration.
Lee et al.~\cite{lee2022extrinsic} estimate camera poses with 3D joints of moving humans.
Recently, HSfM~\cite{mueller2024hsfm} leverages humans to recover the metric scale of camera and scene coordinates obtained from DUSt3R~\cite{dust3r}.
Our work shares the same key insights with this line of research; namely, humans provide powerful cues in challenging scenarios where traditional camera calibration method fails, such as scenes with repetitive structures, textureless scenes, or wide baseline setup.
Also, utilizing humans allows us to recover the metric scale of reconstructed scenes and camera poses, which cannot be achieved with traditional SfM methods.
However, previous methods only work for synchronized videos from static cameras~\cite{lee2022extrinsic} or static scenes~\cite{mueller2024hsfm,xu2021wide,ma2022virtual}.
Moreover, the estimated poses with humans are not accurate enough to directly reconstruct NeRF.

\paragraph{Dynamic 3D scene reconstruction}
The emergence of NeRF~\cite{nerf} and its extension to the temporal domain enables immersive experience in a dynamic world, which was previously limited to a small navigatable area and requires a complicated capturing system~\cite{msi}.
One of the most common approaches to reconstruct 4D scenes is reconstructing 3D scenes in a canonical frame and optimize a deformation field to warp them~\cite{nerfies,hypernerf,dnerf,neuralsceneflow,rodynerf,banmo}.
However, as pointed out by Park et al.~\cite{temp_interp_is_all_you_need}, it is hard to let the deformation network learn general dynamics in practice.
On the other hand, recent grid-based methods~\cite{kplanes,hexplane} simply represent 4D scenes by introducing an additional temporal domain without deformation field, which enables to represent general dynamic scenes easily.
We exploit K-Planes~\cite{kplanes} for our 4D scene representation due to its versatility.

However, all of the aforementioned dynamic NeRFs need synchronized videos and ground-truth camera poses.
Recently, Sync-NeRF~\cite{syncnerf} aims to reconstruct dynamic NeRF from unsynchronized videos.
They additionally optimize time offsets during the dynamic NeRF reconstruction.
However, Sync-NeRF requires initial time offsets close to ground truth since it can only deal with small temporal perturbations and they also require known camera poses.

\paragraph{Camera pose estimation with NeRF}
Since iNeRF~\cite{inerf} has shown that one can optimize camera poses given pretrained NeRF with photometric loss, recent works attempt to train NeRF from unknown camera poses.
They ease the burden of the strict requirements of NeRF (i.e., perfect camera poses) by jointly optimizing camera poses during NeRF training.
Instead of na\"ive optimization~\cite{nerfmm}, recent works propose coarse-to-fine optimization~\cite{barf} and curriculum learning strategy that progressively adds camera parameters~\cite{scnerf} for robust optimization.
However, previous works only optimize camera poses with \emph{static} NeRF.
To the best of our knowledge, we are the first to reconstruct dynamic NeRF from unknown time offsets and camera poses.
We also observe that progressive training is critical for robust optimization in dynamic NeRF scenarios.
We propose a coarse-to-fine optimization technique for multiresolution grid representation and curriculum learning strategy.
More importantly, all previous works need \emph{good initial points} and only adjust small misalignments during joint optimization or only work for image sets with extremely small baselines (e.g., forward-facing capture~\cite{llff}).
We propose a novel way to obtain good initial points with humans, both for time offsets and camera poses.

\section{Method}

\subsection{Problem Setup and Calibration Preparation}
\label{subsec:method_preparation}
\paragraph{Problem setup}
We reconstruct dynamic NeRF from unsynchronized multi-view videos whose poses are not known.
We assume that 1) dynamic scenes contain moving humans (we do not assume that humans are the only dynamic objects), 2) known intrinsics, and 3) known person correspondence across views when there are multiple humans.
Note that we can handle an arbitrary number of humans in the scene.
Also, we impose a minimal assumption on the camera parameters -- we can handle moving cameras with different intrinsic or frame rates. 

Each of $N$ multi-view input video $V^i, i\in[1,N]$ contains $M^i$ image frames, $V^i=(I_t^i; t\in [0,M^i))$, where $I_t$ is image frame at time $t$.
We estimate time offset $\Delta t^i\in\mathbb{R}$ such that the timestamp $t$ for the $i$th video $V^i$ can be mapped to a synchronized global timestamp by adding time offset $t + \Delta t^i$.
Furthermore, we aim to find camera poses in world coordinate, namely rotation $R^i_t$ and translation $\tau^i_t$ of each video.
Hereafter, we will use the term ``calibration parameters'' to refer to both time offsets and camera poses.

\newcommand{\seq}{{=}} 

\paragraph{Human motion estimation}
We first estimate human motions in each multi-view video and use them as a cue to calibrate time offset and camera pose.
We utilize SLAHMR~\cite{slahmr} which is a method to recover 3D human motion from monocular video.
Given a video $V^i$, we extract human motion $\mathbf{H}^i$ which can be expressed as a time series,
\begin{equation}
    \textstyle
    \mathbf{H}^i\seq\left(\bigoplus_{k=1}^{K}h_{k,t}^i;t\in [0,M^i)\right),
\end{equation}
where $K$ is the number of humans in the scene, $\oplus$ is concatenation operator, $h_{k,t}^i\seq\{\Phi_{k,t}^i,\Theta_{k,t}^i,\beta_k^i,\Gamma_{k,t}^i\}$ is state of the $k$th human at time $t$ which is composed of root orientation $\Phi_{k,t}^i\in\mathbb{R}^3$, body pose $\Theta_{k,t}^i\in\mathbb{R}^{22\times 3}$ that is modeled by relative 3D rotation of joints in axis-angle representation, position of root $\Gamma_{k,t}^i\in\mathbb{R}^3$, and human shape parameters $\beta_k^i\in\mathbb{R}^{16}$ which remains constant across whole time.
Human motions $\mathbf{H}^i$s are extracted from each monocular video independently since videos are unsynchronized and we cannot leverage correspondence across different videos.

\begin{figure}
    \centering
    \includegraphics[trim={0, 3mm, 0, 0},clip,width=0.9\linewidth]{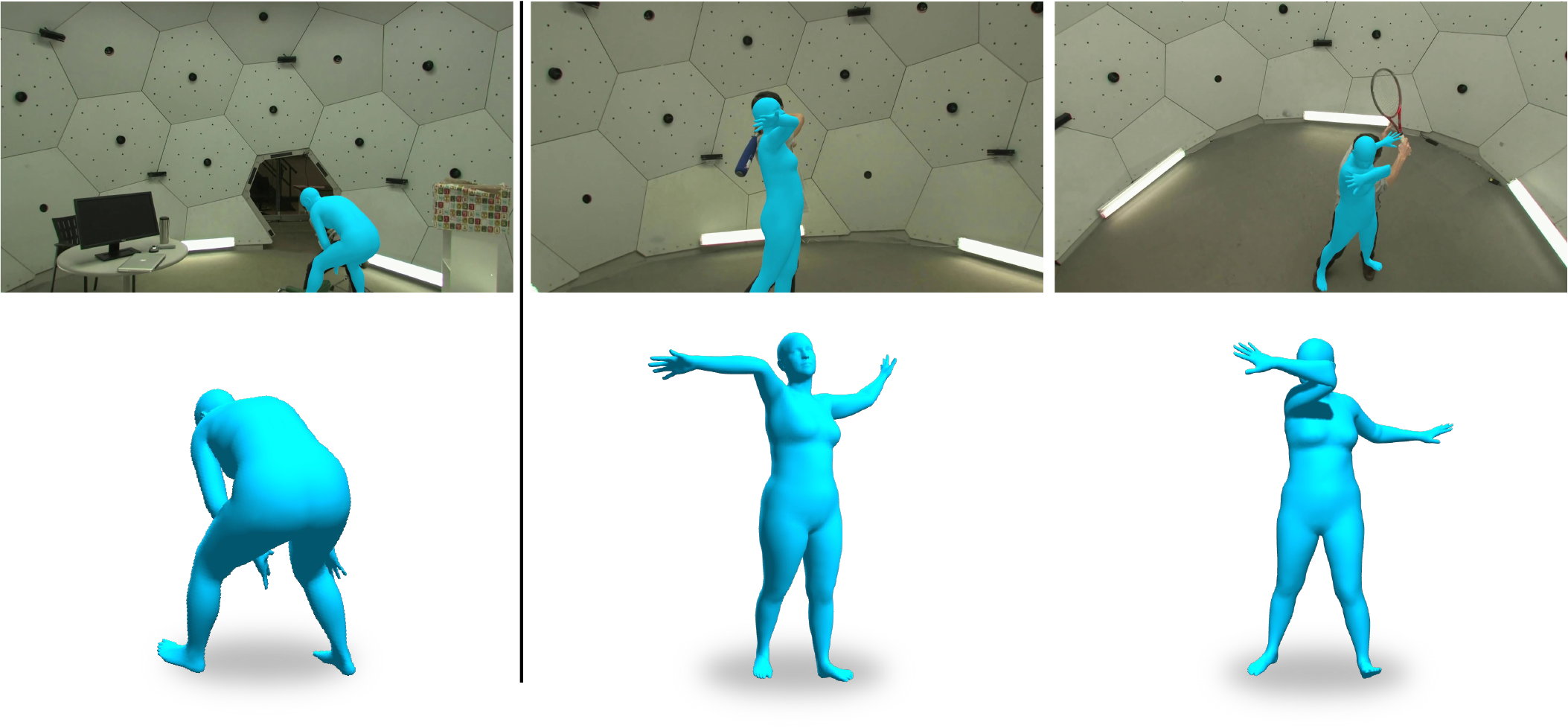}
    \caption{Examples of estimated human motion on our dataset. While SLAHMR~\cite{slahmr} can reconstruct valid human motion for most frames (left), it sometimes fails when there is ambiguity due to the viewing angle (middle) or fast motion (right).}
    \label{fig:slahmr_results}
\end{figure}

We can also obtain camera trajectory relative to human motion.
First, we obtain camera trajectory with SLAM method~\cite{droidslam} whose scale is ambiguous.
Then, we recover scale with data-driven human motion priors during the optimization process following SLAHMR.
For more details, we refer the readers to the original paper~\cite{slahmr}.
We demonstrate estimated human motion in~\cref{fig:slahmr_results}.
It is noteworthy that even if the estimated human poses are not very accurate for some frames, we can robustly estimate calibration parameters with our initialization stage in~\cref{subsec:method_initialization}.

\subsection{Initialization Stage}
\label{subsec:method_initialization}
With extracted human motions $\mathbf{H}$, we estimate time offsets of unsynchronized videos and camera poses in global coordinates.
These estimated calibration parameters serve as initial points for refinement during the optimization of dynamic NeRF afterward (\cref{subsec:method_joint_optimization}).

\paragraph{Time offset estimation}
We first estimate the time offset between a pair of videos, $V^i$ and $V^j$.
We consider time offset estimation as an alignment problem between time series of human motion $\mathbf{H}$.
To do so, we define matching cost, or distance between arbitrary two states $h^i_{t_1}$ and $h^j_{t_2}$ from motion sequence $\mathbf{H}^i$ and $\mathbf{H}^j$, respectively:
\begin{equation}
    d(h^i_{t_1}, h^j_{t_2}) = \lVert \mathbf{J}_{\text{canon},t_1}^i - \mathbf{J}_{\text{canon},t_2}^j\rVert_2,
    \label{eq:human_dist}
\end{equation}
where $\mathbf{J}_{\text{canon}}\in\mathbb{R}^{22\times 3}$ is a 3D positions of human body joints relative to the root. 
Joint positions can be obtained with SMPL-H~\cite{smplh} model $\mathcal{S}$, which takes root rotation, body pose, and shape parameter as input.
We set root rotation as $\mathbf 0$ to get joint positions at canonical frame,
\begin{equation}
    \mathbf{J}_{\text{canon},t}^i = \mathcal{S}_{\text{canon}}(h_t^i) = \mathcal{S}(\mathbf{0}, \Theta_t^i,\beta^i).
\end{equation}
We use the explicit 3D joint positions for the loss instead of comparing distances directly from human pose parameter $\Theta_t$ and shape parameter $\beta$.
The loss in the physical ambient space better reflects the actual deviation.

We find all pairwise alignments of human motions and store them into two $N\times N$ matrices. 
Each pairwise alignment employs a dynamic time warping (DTW) algorithm, which is widely used in speech recognition~\cite{dtw_speech}.
\begin{equation}
    C_{ij}, \Delta T_{ij} = \text{DTW}(\mathbf{H}^i, \mathbf{H}^j), \forall i<j,
\end{equation}
where time offset matrix $\Delta T\in\mathbb{Z}^{N\times N}$ stores the estimated relative time offsets and cost matrix $C\in\mathbb{R}^{N\times N}$ stores the cost of the alignment at the estimated time offset.
To calculate the cost with the DTW algorithm, we use the distance between human states defined in~\cref{eq:human_dist}.
We estimate time offset $\Delta{T}_{ij}$ as the most frequent warping time.
If the framerate of two videos are different, we upsample 3D joint locations by linearly interpolating joint locations between observed frames of lower framerate videos.

Then, we find the global sequence alignment of whole human motions with a greedy algorithm.
First, we find an anchor index pair which has the minimum value $(i,j)=\argmin_{(i,j)}C_{ij}$.
Then, we incrementally add index pairs in increasing order with respect to the cost value
\begin{equation}
    \Delta t = \text{\textsc{Global Align}}(C, \Delta T),
\end{equation}
where $\Delta t \in \mathbb{Z}^N$ is estimated time offsets that globally align human motions.
A detailed pseudocode for the global alignment process is provided in the supplemental material.

\paragraph{Camera pose estimation}
Next, we estimate camera poses by aligning human motions after the initial time offset estimation.
Similar to the time offset estimation, we utilize 3D joint positions as calibration patterns.
Different from time alignment, we extract joint positions by considering the orientation and translation of human roots:
\begin{equation}
    \mathbf{J}_{\text{global},t}^i = \mathcal{S}_{\text{global}}(h_t^i) = \mathcal{S}(\Phi_t^i, \Theta_t^i,\beta^i) + \Gamma_t^i.
\end{equation}
Then we find the optimal similarity transform $\mathcal{T}^i\in\text{SIM}(3)$ that minimizes the Procrustes distance between human motion $\mathbf{H}^i$ and anchor human motion $\mathbf{H}^{\alpha}$ for all index $i$ except the anchor index $\alpha$,
\begin{equation}
    \mathcal{T}^i\seq \text{{Procrustes}}\left((\mathbf{J}_{\text{global},t+\Delta t^{\alpha}}^{\alpha};t), (\mathbf{J}_{\text{global},t+\Delta{t}^i}^i;t)\right),
\end{equation}
where anchor index $\alpha$ is randomly selected ($\alpha\in[1,N]$).
We describe details of Procrustes analysis in the supplementary material.

Then we can obtain camera poses $R^i_t, \tau^i_t$ of the $i$th camera by applying the same similarity transform $\mathcal{T}^i$ to camera poses that are obtained with human motions in~\cref{subsec:method_preparation}.

\subsection{Refinement with Dynamic NeRF Optimization}
\label{subsec:method_joint_optimization}

In this step, we further refine time offsets and camera poses, which are initialized properly in the initialization stage (\cref{subsec:method_initialization}) during the optimization of dynamic NeRF.
We utilize K-Planes~\cite{kplanes} as our dynamic NeRF representation.
K-Planes uses six multi-resolution feature grids for 4D NeRF.
Namely, there are $\mathbf{P}^{xy}_l$, $\mathbf{P}^{yz}_l$, and $\mathbf{P}^{zx}_l$ for space-only grids, $\mathbf{P}^{xt}_l$, $\mathbf{P}^{yt}_l$, and $\mathbf{P}^{zt}_l$ for space-time grids, where $l\in[1,L]$ denotes the resolution level.
We can obtain the features at querying spatiotemporal point $\mathbf{q}=(\mathbf{x},t)$ by
\begin{equation}
    \textstyle
    f_l(\mathbf{q})=\bigodot_{c}\psi(\mathbf{P}^c_l,\pi^c(\mathbf{q})),
\end{equation}
where $\bigodot$ is a Hadamard product, $\pi^c$ is the projection operator that maps $\mathbf{q}$ onto the $c$th plane, and $\psi$ is a bilinear interpolation.
Then, multi-level features are concatenated to a single feature vector $f(\mathbf{q})$
\begin{equation}
    \textstyle
    f(\mathbf{q}) = \bigoplus_{l=1}^L w_lf_l(\mathbf{q}),
\label{eq:feat_concat}
\end{equation}
where $w_l$ is a weight multiplied to $l$-level feature vector.
Volume density and color at spatiotemporal point $\mathbf{q}$ with viewing direction $\mathbf{d}$ can be obtained by
\begin{gather}
    \sigma(\mathbf{q}), \hat{f}(\mathbf{q}) = \mathcal{F}_{\sigma}(f(\mathbf{q})),\\
    \mathbf{c}(\mathbf{q},\mathbf{d}) = \mathcal{F}_{\text{color}}(\hat{f}(\mathbf{q}),\gamma(\mathbf{d})),
\end{gather}
where $\mathcal{F}_{\sigma}$ and $\mathcal{F}_{\text{color}}$ are tiny MLPs and $\gamma$ is a positional encoder~\cite{nerf}.
Then, we can obtain the pixel value at time $t$ along a ray $\mathbf{r}_t^i=(\mathbf{o}_t^i,\mathbf{d}_t^i)$ with volume rendering formulation.
$\mathbf{o}_t^i$ and $\mathbf{d}_t^i$ are the camera center and ray direction, which can be obtained with $i$th camera pose $R_t^i, \tau_t^i$
\begin{equation}
    \textstyle
    \hat{I}\seq\sum_{k=1}^N T_k(1-e^{\sigma(\mathbf{x}_k, t+\Delta t^i)\delta_k})\mathbf{c}((\mathbf{x}_k,t+\Delta t^i), \mathbf{d}_t^i),
\end{equation}
where $\mathbf{x}_k=\mathbf{o}_t^i+p_k\mathbf{d}_t^i, p_k\in[\text{near},\text{far}]$ is the sample point on ray, $T_k=e^{-\sum_{j=1}^{k-1}\sigma_j\delta_j}$ is the accumulated transmittance, and $\delta$ is the distance between ray samples.

We optimize both 4D radiance fields and calibration parameters by minimizing photometric loss
\begin{equation}
    \textstyle
    \mathcal{L}=\sum_{i,t,\mathbf{r}}\lVert I_t^i(\mathbf{r}) - \hat{I}(\mathbf{r}_t^i,t+\Delta t^i) \rVert_2,
\label{eq:rendering_loss}
\end{equation}
where $I_t^i(\mathbf{r})$ is a ground-truth pixel corresponding to randomly selected ray $\mathbf{r}$ of $t$th frame of $i$th camera.

However, na\"ive joint optimization of dynamic NeRF and calibration parameters easily falls into bad local minima, although we have good initial points for camera poses and time offsets.
We propose a progressive learning strategy for robust joint optimization.

First, we observe that coarse-to-fine registration is critical for dynamic NeRF.
We weigh the feature from resolution level $l$ by $w_l$ before concatenated to single feature vector at~\cref{eq:feat_concat}:
\small
\begin{equation}
    w_l=\begin{cases}
        0 & \text{if } \alpha < l-1,\\
        \frac{1-\cos((\alpha-(l-1))\pi)}{2} & \text{if } 0\leq \alpha-(l-1)<1\\
        1 & \text{if } \alpha - (l-1) \geq 1
    \end{cases},
\end{equation}
\normalsize
where $\alpha=L(e^{\eta}-1)/(e-1)$, $\eta\in[0,1]$ is normalized training step.
This is in contrast to vanilla K-Planes~\cite{kplanes}, which uses all weights as 1.
Our feature weighting strategy assigns big weights to low-frequency grids at the beginning of the optimization stage and progressively increases weights of higher-resolution grids.
Our strategy is inspired by the dynamic low-pass filter in BARF~\cite{barf}, but we instead modify the relative weights for grid resolution to effectively start from low-frequency features in multi-resolution K-Planes.

Furthermore, we propose a curriculum learning strategy for stable optimization.
We first freeze camera poses and time offsets and only optimize 4D NeRF for $s_0$ training steps.
Then, we add camera poses to the parameter list from $s_0$ iterations.
Finally, we add time offsets to learnable parameters and jointly optimize all parameters after another $s_1$ iterations.
This progressive learning strategy is critical in preventing the model from overfitting.
This also aligns with observations from SCNeRF~\cite{scnerf} that sequentially adds complexity to the camera model.

\section{Experiments}

\subsection{Experimental Setup}
\label{subsec:experimental_setup}
\paragraph{Dataset}
We evaluate our method on \emph{real-world} datasets containing multi-view videos of dynamic scenes with human motions, including CMU Panoptic Studio~\cite{panoptic}, Mobile-Stage~\cite{4k4d}, and EgoBody~\cite{egobody}.
We take subsequences of each multi-view video that start from random global timestamps, but all of them have overlapping frames.
The ground truth time offsets and camera poses are used for evaluation and not provided to the system.
CMU Panoptic Studio~\cite{panoptic} contains various human motions, interactions with objects, and a large portion of occlusions for some viewpoints, which are captured from $31$ cameras located in the upper hemisphere surrounding the scene.
We use $29$ or $30$ cameras for training dynamic NeRF and one camera for testing novel-view synthesis performance.
Mobile-Stage dataset from 4K4D~\cite{4k4d} contains three dancers captured by smartphone cameras with frontal and side views.
We use $20$ cameras for training and one camera for testing.
EgoBody~\cite{egobody} contains 4 or 5 static Kinect cameras and one moving head-mounted HoloLens2 camera, which has different intrinsics and framerate to Kinect.
Since reconstructing NeRFs from sparse views (6 views) is challenging and an open problem, we focus on the evaluation of initialization stage performance on the EgoBody dataset.
More details on datasets can be found in the supplemental material.

\paragraph{Implementation details}
We optimize K-Planes with Adam optimizer~\cite{adam} with rendering loss in~\cref{eq:rendering_loss} and regularization losses, including total variance in space, time smoothness loss, density L1 loss, and sparse transient loss introduced in original K-Planes~\cite{kplanes}.
We also use proposal sampling~\cite{mipnerf360} to sample ray points for volume rendering.
We jointly optimize calibration parameters and K-Planes for a total of 300k iterations and finish coarse-to-fine scheduling at 100k iterations.
We use a grid resolution of 240 for the time axis and a multi-scale level of $L=4$ or $5$ depending on scenes at the maximum resolution of 384.
We unfreeze camera pose parameters at $s_0=2$k steps, and time offset parameters at $s_0+s_1=20$k steps.
\subsection{Alignment Performance}
\label{subsec:evaluation}
\begin{figure}
    \centering
    \includegraphics[trim={5mm, 5mm, 0, 0}, clip, width=\linewidth]{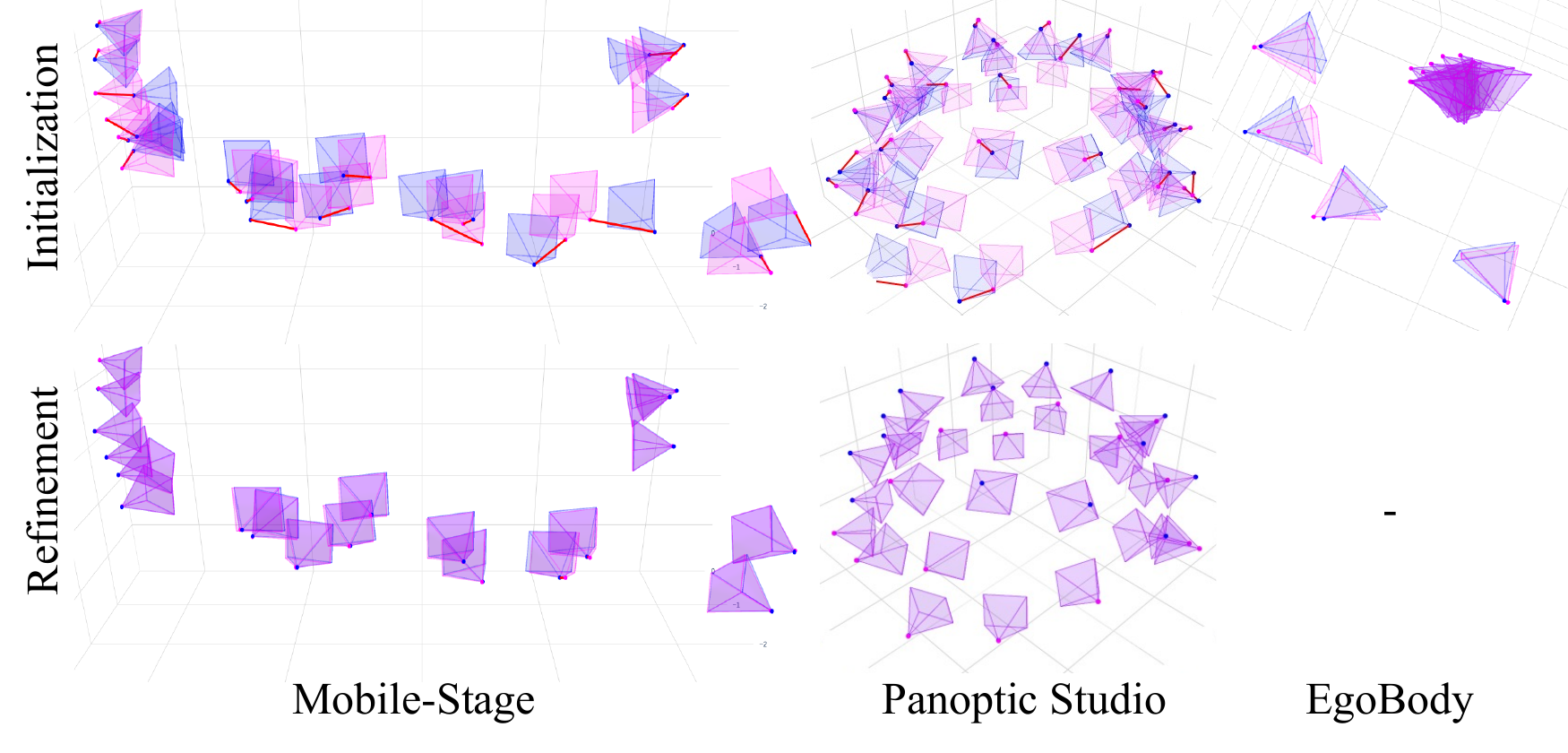}
    \caption{We visualize estimated camera poses of the initialization step (top) and after refinement (bottom) on various datasets. Blue and red frustums are estimation and ground truth, respectively.}
    \label{fig:pose_align}
\end{figure}
\begin{table}
    \centering
    \resizebox{\linewidth}{!}{
    \begin{tabular}{@{}l|l|cc|cc|ccc@{}}
    \toprule
    \multirow{2}{*}{Dataset} & \multirow{2}{*}{Scene} & \multicolumn{2}{c|}{Rotation ($^\circ$)} & \multicolumn{2}{c|}{Trans.}& \multicolumn{3}{c}{$\Delta t$ (frames)}\\
    & & Init & Refine & Init & Refine & Data & Init & Refine\\
    \midrule
    & \textsc{Baseball} & 5.30 & 0.328 & 22.6 cm & 0.16 cm & 29.04 & 0.700 & 0.025 \\
    & \textsc{Office1} & 3.88 & 0.420 & 19.7 cm & 0.17 cm & 33.25 & 3.448 & 0.031 \\
    Panoptic& \textsc{Office2} & 8.89 & 0.660 & 38.0 cm & 0.37 cm & 29.65 & 1.300 & 0.029\\
    Studio& \textsc{Office3} & 3.95 & 0.363 & 19.2 cm & 0.17 cm & 32.93 & 0.800 & 0.026 \\
    & \textsc{Tennis} & 5.29 & 0.290 & 25.8 cm & 0.22 cm & 28.33 & 0.467 & 0.028 \\
     \cmidrule{2-9}
    & Average & 5.46 & 0.412 & 25.1 cm & 0.22 cm & 30.64 & 1.343 & 0.028 \\
    \midrule
    Mobile-Stage & \textsc{Dance} & 5.65 & 1.707 & 0.053 & 0.002 & 24.76 & 0.455 & 0.214 \\
    \midrule
    \multirow{4}{*}{EgoBody} &S22-S21-02 & 12.7 & - & 0.016 & - & 21.94 & 7.200 & - \\
     & S32-S31-01 & 3.68 & - & 0.025 & - & 33.30 & 0.333 & - \\
     & S32-S31-02 & 10.8 & - & 0.034 & - & 8.333 & 1.667 & - \\
     \cmidrule{2-9}
     & Average & 9.07 & - & 0.025 & - & 21.19 & 3.067 & - \\
    \bottomrule
    \end{tabular}
    }
    \caption{Quantitative results on camera pose estimation and time synchronization. We report the results from the first initialization stage (Init) and the second refinement stage (Refine). We set the size of scene bounding box as 1 to measure translation error for Mobile-Stage and EgoBody since the exact metric is not known.}
    \label{tab:align_accuracy}
\end{table}

First, we report the accuracy of time synchronization and camera pose estimation in~\cref{tab:align_accuracy}.
We measure errors after aligning estimated results to ground truth to recover original scale, orientation, and shifts.
Both quantitative results in~\cref{tab:align_accuracy} (``Init'' columns) and visual illustration in~\cref{fig:pose_align} (top) show that our initialization step can achieve reasonable first estimation across all datasets.
Our initialization stage can robustly estimate calibration parameters regardless of the number of humans in the scene (one in Panoptic Studio, two in EgoBody, and three in Mobile-Stage), on various camera configurations (uniformly distributed in Panoptic Studio, sparse distributed in frontal and side views in Mobile-Stage, including moving camera in EgoBody).
Our first alignment with human motion achieves camera pose calibration with a rotation error of less than $5.5^\circ$ and a translation error of less than $26$ cm on average in Panoptic Studio, despite the absence of any input camera pose information.
Furthermore, our initialization strategy can estimate fairly accurate time offsets (average 1.34 frames) even from severely unsynchronized videos (average 30.64 frames offset).
Also, our initialization stage consistently estimates accurate calibration parameters in Mobile-Stage and EgoBody dataset.

Starting from the initial estimation, our refinement with joint optimization of 4D NeRF can achieve near-perfect calibration as shown in~\cref{tab:align_accuracy} (``Refine'' columns) and~\cref{fig:pose_align} (bottom).
Our approach aligns camera poses within rotation error of $0.4^\circ$ and translation error of $0.22$ cm on average in Panoptic Studio.
Also, the error of estimated time offsets is less than 0.03 frames on average, which is equal to 1 millisecond.
Our method also shows highly accurate calibration after refinement in Mobile-Stage dataset as well.

We also test to use different distance function, or matching cost between two human states.
Instead of L2 distance between 3D joint positions as defined in~\cref{eq:human_dist}, we test na\"ive L2 distance between human shape and pose parameters for time offset estimation at initialization step:
\small
\begin{equation}
    d(h_{t_1}^i,h_{t_2}^j)=\lVert [\Theta_{t_1}^i,\beta^i] - [\Theta_{t_2}^j,\beta^j]\rVert_2.
\end{equation}
\normalsize
As shown in~\cref{tab:robustness} (first row), replacing the distance function degrades the time synchronization performance significantly considering we are using the same SMPL sequence.

\begin{figure}
    \centering
    \includegraphics[trim={0, 5mm, 0, 0}, clip, width=0.85\linewidth]{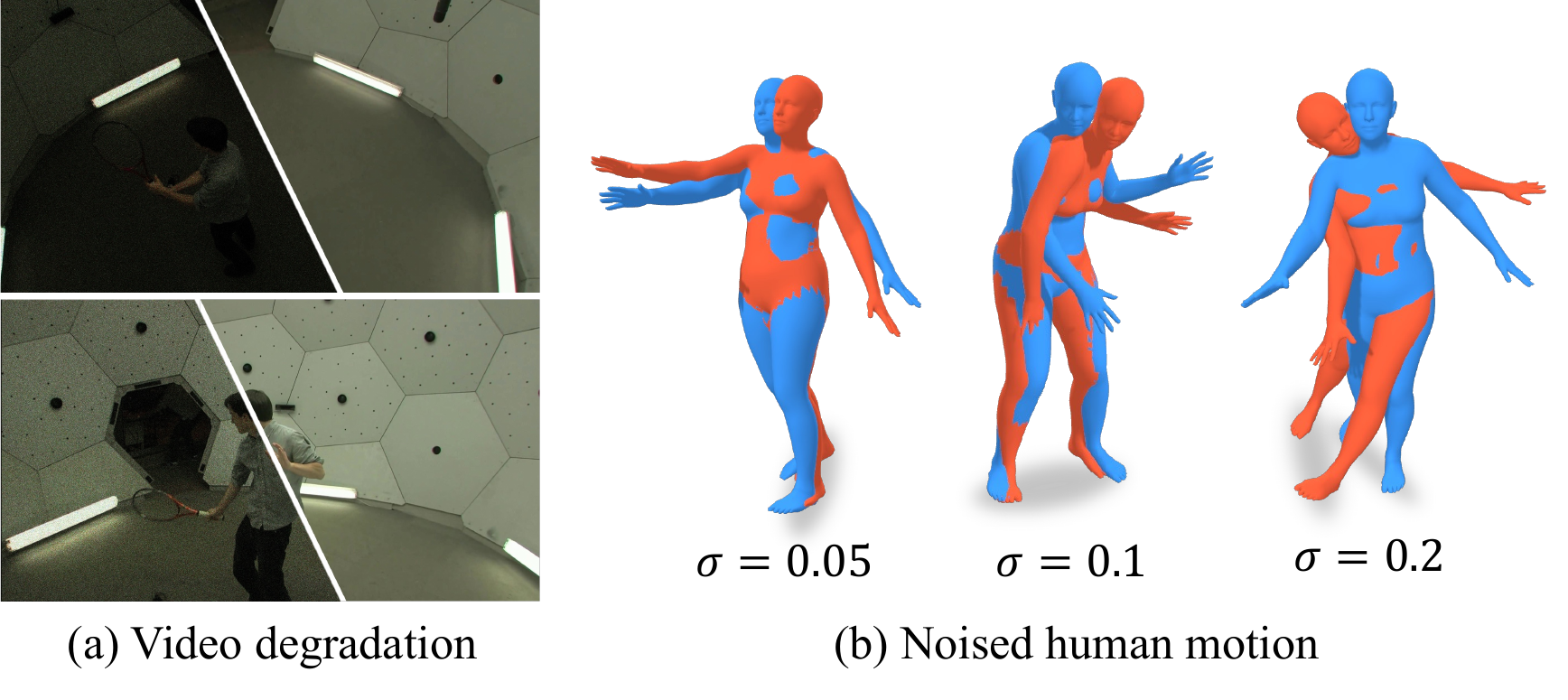}
    \caption{Visual examples of (a) video degradation and (b) perturbed human motion with Gaussian noise of various $\sigma$ to test robustness of our method. Blue human meshes are initially estimated human motion and red meshes are noised results.}
    \label{fig:noise_example}
    \vspace{-1em}
\end{figure}

We further investigate the robustness of our initialization stage.
First, we evaluate our method after applying image degradation and color variations to each training video $V^i$.
We apply gamma correction with randomly sampled within uniform range $\gamma\sim[0.35,2.1]$ for color variations and add luminance noise and chromatic noise for image degradation.
We also add noise to human shape parameter $\beta^i$ and pose parameter $\Theta_t^i$ to test robustness on the accuracy of SMPL parameter estimation.
Samples of degraded images and noised human motion can be found in~\cref{fig:noise_example}.
Additionally, we test mixed-framerate input using FPS=24 for five videos and FPS=30 for others.

\begin{table}
    \centering
    \resizebox{\linewidth}{!}{
    \begin{tabular}{l|cc|cc|cc}
    \toprule
    \multirow{2}{*}{\textsc{Tennis} scene} & \multicolumn{2}{c|}{Rotation ($^\circ$)} & \multicolumn{2}{c|}{Trans. (cm)} & \multicolumn{2}{c}{$\Delta t$ (frames)} \\
    & Init & Refine & Init & Refine & Init & Refine \\
    \midrule
    L2 norm, $\beta, \Theta$ & 5.382 & 0.656 & 26.130 & 0.261 & 2.100  & 0.030 \\
    SMPL noise, $\sigma=0.01$ & 5.266 & 0.285  & 25.844 & 0.296 & 1.633 & 0.027 \\
    SMPL noise, $\sigma=0.02$ & 5.307 & 1.240  & 25.825 & 0.276  & 0.533 & 0.028  \\
    SMPL noise, $\sigma=0.05$ & 5.310 & 0.357   & 25.861 & 0.277  & 0.833 & 0.029  \\
    SMPL noise, $\sigma=0.1$ & 5.241 & 0.518  & 25.383 & 0.257  & 2.100 & 0.030  \\
    SMPL noise, $\sigma=0.2$ & 5.404 & 0.934  & 26.831 & 0.217  & 2.367 & 0.024 \\
    Degraded video & 6.496 & - & 32.374 & - & 1.133 & -\\
    Mixed FPS & 7.933 & - & 38.58 & - & 1.200 & - \\
    \midrule
    Default setup & 5.293 & 0.290 & 25.827 & 0.217  & 0.467 & 0.028  \\
    \bottomrule
    \end{tabular}
    }
    \caption{Quantitative results with different choice of human state distance, noised motion, input degradation, and mixed FPS setup.}
    \vspace{-0.5em}
    \label{tab:robustness}
\end{table}


\begin{figure}
    \centering
    \includegraphics[trim={0, 3mm, 0, 0}, clip, width=\linewidth]{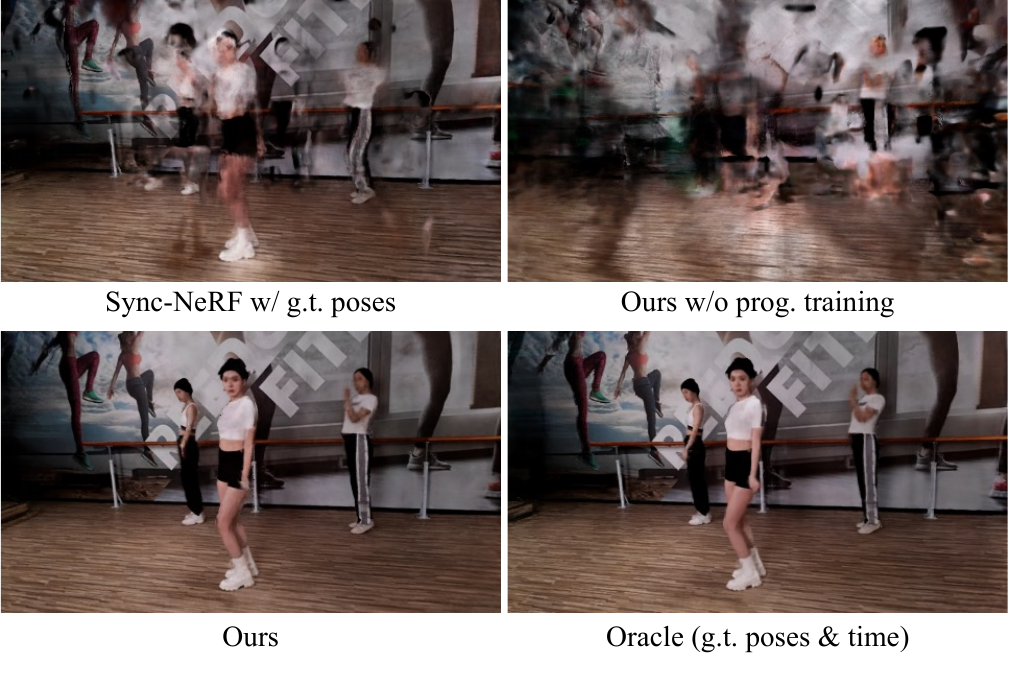}
    \caption{Qualitative comparison on Mobile-Stage dataset}
    \label{fig:qualitative_mobilestage}
\end{figure}

We report quantitative results on robustness experiments in~\cref{tab:robustness}.
Although the alignment performance with degraded videos underperforms compared to the original setup, our approach still produces reasonable time offsets and camera poses for initialization, considering the severe appearance variation and quality degradation.
This result supports our claim that human motion can serve as a robust mid-level representation for calibration.
Our method also estimates reasonable calibration parameters from videos with mixed framerates.
Our approach also shows comparable accuracy to the original data when the noise level is moderate ($\sigma\leq0.05$).
These results show that our aligning strategy is not sensitive to the quality of extracted human motion.
Although time offset errors increase at severe noise levels ($\sigma\geq0.1$), our approach drastically reduces temporal misalignment (from 28.33 of input data to less than 2.5).
Furthermore, calibration parameters initialized from noised human motions were sufficiently accurate to recover precise time offsets and camera poses in the subsequent refinement step, as shown in~\cref{tab:robustness}.

\begin{figure*}
    \centering
    \includegraphics[trim={0, 0, 0, 2mm}, clip, width=\linewidth]{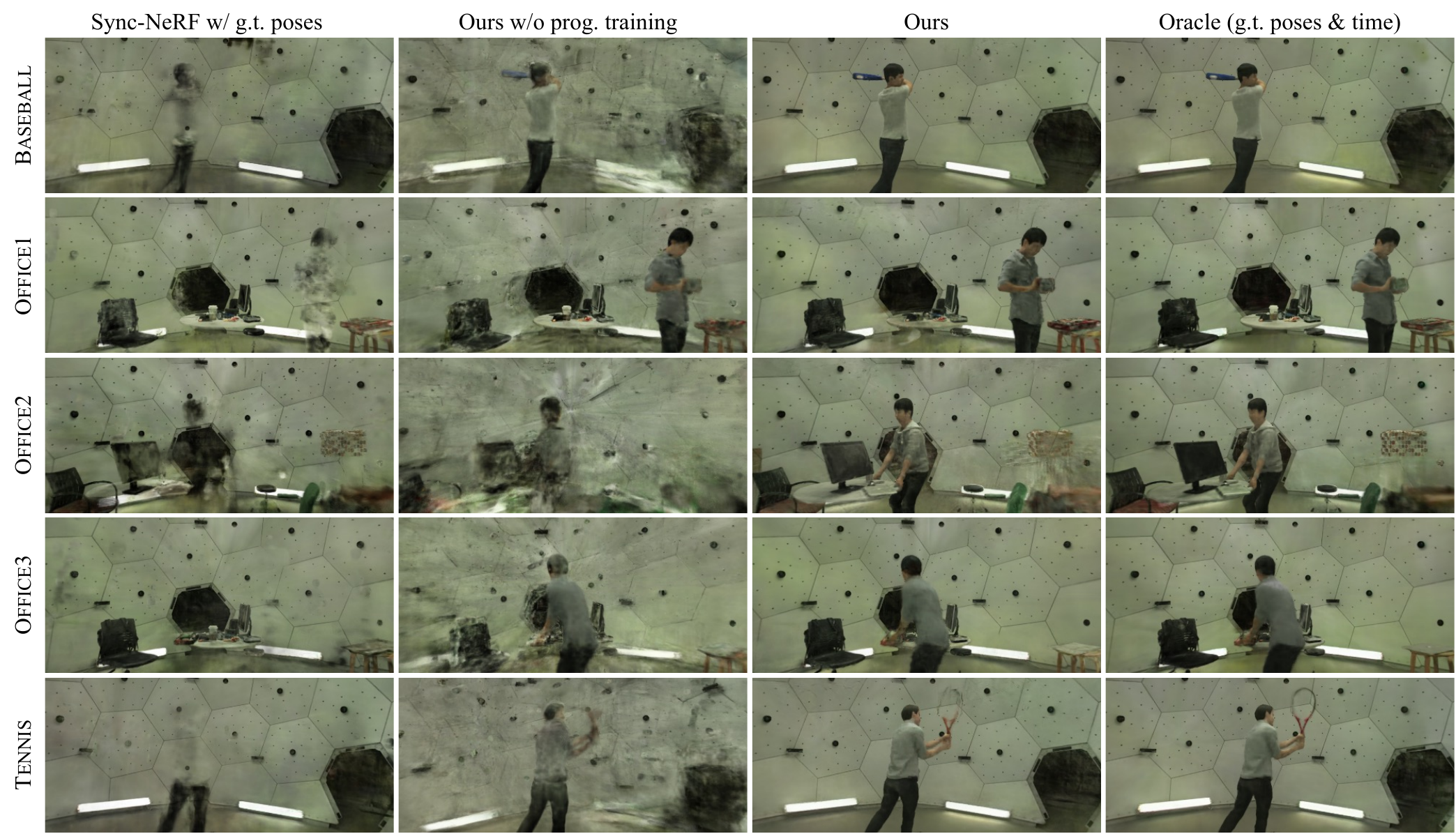}
    \caption{Qualitative comparison of novel view synthesis performance on CMU Panoptic Studio dataset.}
    \label{fig:qualitative_comparison}
\end{figure*}

\begin{table*}
    \centering
    \resizebox{0.8\linewidth}{!}{
    \begin{tabular}{l|l|ccc|ccc|ccc|ccc}
    \toprule
   \multirow{2}{*}{Dataset} & \multirow{2}{*}{Scene} & \multicolumn{3}{c|}{Sync-NeRF w/ GT pose}  & \multicolumn{3}{c|}{Ours w/o prog. training} & \multicolumn{3}{c|}{Ours} & \multicolumn{3}{c}{Oracle (GT pose \& time)}\\
    & & PSNR & SSIM & LPIPS & PSNR & SSIM & LPIPS & PSNR & SSIM & LPIPS & PSNR & SSIM & LPIPS\\
    \midrule
    \multirow{6}{*}{Panoptic Studio} & \textsc{Baseball} & 21.17 & 0.833 & 0.189 & 19.23 & 0.577 & 0.412& 27.20 &0.919 & 0.072  & 26.80 & 0.925 & 0.065\\
    & \textsc{Office1} & 20.62 & 0.809 & 0.196  & 22.33 & 0.678 & 0.334& 26.65 & 0.855 & 0.125  & 27.00 & 0.905 & 0.079\\
    & \textsc{Office2} & 21.14 & 0.782 & 0.195  & 18.39 & 0.511 & 0.544& 24.43 & 0.820 & 0.155 & 26.58 & 0.891 & 0.091\\
    & \textsc{Office3} & 20.94  & 0.826 & 0.178  & 21.06 & 0.614 & 0.383 & 27.51 & 0.893 & 0.093   & 28.23 & 0.912 & 0.077\\
    & \textsc{Tennis} & 22.09 & 0.851 & 0.168 & 17.29 & 0.531 & 0.552& 26.94 &0.881 & 0.113 & 27.22 & 0.916 & 0.080\\
    \cmidrule{2-14}
    & Average & 21.19 & 0.820 & 0.185 &19.66 & 0.582 & 0.445 & 26.55 & 0.874 & 0.112 & 27.16 & 0.910 & 0.078 \\
    \midrule
    {Mobile-Stage} & \textsc{Dance} & 17.26 & 0.410 & 0.284 & 13.72 & 0.233 & 0.510 & 23.27	& 0.816  & 0.089  & 22.98 & 0.806 & 0.093 \\
    \bottomrule
    \end{tabular}
    }
    \caption{Quantitative comparison of novel view synthesis performance.}
    \label{tab:view_synthesis}
\end{table*}

\subsection{Dynamic Novel-view Synthesis}

To evaluate our dynamic scene reconstruction performance, we measure novel view synthesis errors.
We measure photometric error for rendered images at test view only for timestamps that are overlapped by all of the training videos since we take unsynchronized videos as input.
We also conduct test-time optimization for accurate measurement that freezes NeRF parameters and optimizes only test camera poses and timestamps for small iterations before measuring metrics.
Since we are the first approach to reconstruct dynamic NeRF without any assumption of both camera poses and time synchronization, there is no existing work to compare with our method that has exactly the same problem setup.
Instead, we compare our approach to Sync-NeRF~\cite{syncnerf}, which is a method that jointly optimizes time offsets with dynamic NeRF.
Since Sync-NeRF cannot align camera poses, we relax our problem setup for Sync-NeRF by providing ground-truth camera poses.
We also compare our method with the oracle, namely, K-Planes with ground-truth camera poses and time offsets provided by the dataset.

Quantitative results of novel view synthesis in mean PSNR, SSIM~\cite{ssim}, and LPIPS~\cite{lpips} are reported in~\cref{tab:view_synthesis}.
Our approach achieves superior performance across all metrics compared to Sync-NeRF even if Sync-NeRF baseline uses ground-truth camera poses.
It supports our claim that a good initial point (time offset for Sync-NeRF comparison) is critical for gradient-based optimization.
Without proper initialization of time offsets, Sync-NeRF cannot reconstruct high-quality dynamic scenes.
Furthermore, our approach achieves almost on par performance to the oracle model.
Our approach achieves a slightly higher PSNR than the oracle in the \textsc{Baseball} scene in Panoptic Studio and Mobile-Stage.
This is possible because the ground-truth camera poses in the real-world dataset are not perfectly accurate, and our method optimizes time offsets to subframe precision.
We show qualitative results rendered at novel viewpoints in~\cref{fig:qualitative_comparison,fig:qualitative_mobilestage}.
Our approach achieves superior rendering quality compared to Sync-NeRF baseline and comparable results to the oracle model, which aligns with the quantitative results.
Sync-NeRF cannot calibrate time offsets without proper initialization and cannot reconstruct dynamic objects consequently.

Additionally, we test our approach without our progressive learning strategy with initialized calibration parameters. 
Quantitative results and qualitative results are also included in~\cref{tab:view_synthesis} and~\cref{fig:qualitative_comparison,fig:qualitative_mobilestage}, respectively.
Even starting with proper camera poses and time offsets estimated by our initialization stage, without progressive training, the reconstructed dynamic scenes show inferior quality.
This result validates the importance of the proposed progressive learning strategy.

\section{Conclusion}
We propose a practical solution to reconstruct neural dynamic 3D scenes containing humans from unsynchronized and uncalibrated multi-view videos.
Given human motion extracted from individual videos, we find the initial estimates of temporal offsets and camera poses, which are highly robust to occlusions or other adversaries.
We then train 4D NeRF volume in a coarse-to-fine fashion, which effectively stabilizes the optimization process.
During training, we progressively add the estimated parameters into a joint optimization routine and further refine them.
We demonstrate that we can acquire reliable dynamic scene reconstruction in challenging setups performing on par with accurate calibration.

\noindent
\textbf{Limitations \& future works}
Although our initialization stage robustly estimates time offsets and camera poses for most cases, thanks to the temporal aggregation, we cannot recover calibration parameters if SLAHMR~\cite{slahmr} completely fails to recover human motion from video.
Nevertheless, we anticipate that as advancements in human motion estimation continue, our approach will benefit from these improvements.
Our formulation is general enough to handle moving cameras; however, refining camera poses of moving cameras by joint optimization with 4D NeRF is left as future work.
Additionally, integrating techniques that address varying appearances and occlusions (e.g., appearance embedding and uncertainty estimation~\cite{nerfinthewild}) could enable our method to reconstruct complex scenes from distributed videos, such as concert halls or sports games, directly from fan-captured internet videos.
Also, extending our work beyond humans to generalize to other common objects (e.g., cars) would be also an exciting direction.

{
    \small
    \bibliographystyle{ieeenat_fullname}
    \bibliography{main}
}

\clearpage
\setcounter{page}{1}
\maketitlesupplementary
\appendix
\section{Dataset Details}
\paragraph{CMU Panoptic Studio}
CMU Panoptic Studio dataset contains dynamic sequences of human movement from 31 cameras surrounding the scene.
We take subsequences from sports and office sequence and make new dataset containing five scenes \textsc{Baseball, Tennis, Office1, Office2}, and \textsc{Office3}.
There is one human in the scenes in Panoptic Studio dataset we used.
Each multi-view training video is 270 frames long at 30 FPS and starts from a random global timestamp to make unsynchronized setup.
All video sequences are sampled to have at least 150 overlapping frames.
Namely, maximum time offset between two videos is 120 frames.
We undistort all training images before estimating human motion and training dynamic NeRFs with provided radial and tangential distortion parameters.

\paragraph{Mobile-Stage}
Mobile-Stage dataset from 4K4D contains three dancers captured by multiple smartphone cameras with frontal and side views.
Some viewpoints have significant occlusion, and not all people are visible in certain viewpoints.
The video lengths, FPS, and random global timestamp sampling strategy are identical to the setup in CMU Panoptic Studio dataset.
We use 20 cameras for training and one camera for evaluation.

\paragraph{EgoBody}
EgoBody dataset contains dynamic interaction between two humans that are captured from four (S22-S21-02) or five (S32-S31-01 and S32-S31-02) static Kinect cameras and one moving head-mounted HoloLens2 camera.
HoloLens2 camera has different camera intrinsic from Kinect cameras and it has some missing frames.
Also, there are extreme motion blurred frames in HoloLens2 camera videos.
We estimate human pose for existing frame with SLAHMR and estimate human pose for missing frames by linear interpolation of 3D joint positions.
Each multi-view training video is 200 frames long.
They have at least 90 overlapping frames, in other words, maximum time offset between two videos is 110 frames.

\section{Implementation Details}
\subsection{Training K-Planes}
\label{subsec:training_kplanes}
We use $L$=5 spatial grid resolutions $[24, 48, 96, 192, 384]$ for Mobile-Stage dataset and \textsc{Office1, Office2, Office3,} and \textsc{Tennis} scenes of CMU Panoptic Studio dataset, and we use $L$=4 grid resolutions $[48, 96, 192, 384]$ for \textsc{Baseball} scene of CMU Panoptic Studio dataset.
We observe that \textsc{Baseball} scene converges well starting from the resolution of $48$.
We use a single resolution, $240$, for the temporal grid in all scenes.

In addition to the weight scheduling described in the main manuscript, we also schedule the weights of regularization terms.
We apply cosine scheduling that decreases weights to $1/100$ of its initial weights at the end of the scheduling.
We use weights $0.01$ for distortion loss, $0.001$ for L1 loss in time planes, $0.001$ for total variance loss in spatial planes, $0.01$ for time smoothness loss, and $0.01$ for density L1 loss.
We start scheduling of regularization from $100$k steps for Mobile-Stage dataset and \textsc{Office1, Office2, Office3,} and \textsc{Tennis} scenes of Panoptic Studio dataset, and $50$k steps for \textsc{Baseball} scene of Panoptic Studio dataset, and end scheduling at $150$k steps.

For efficiency, we initialize feature values of finer grids by bilinear interpolation of values from coarser grids.
Namely, we initialize finer grids $\mathbf{P}_l^c, (l>1)$ at $\alpha=l-1$ with values interpolated from $\mathbf{P}_{l-1}^c$, where $\alpha=L(e^\eta -1)/(e-1)$, $\eta\in[0,1]$ is a normalized training step.

\subsection{Global Alignment Algorithm}
\label{subsec:global_alignment}

\DontPrintSemicolon
\SetCommentSty{algcommentfont}
\SetKwFunction{KwFn}{Fn}
\SetKw{Continue}{continue}
\SetNlSty{}{}{}

\begin{algorithm}[t]
    \footnotesize
    \SetKwInOut{Input}{Input}
    \SetKwInOut{Output}{Output}
    \SetKwProg{Func}{Function}{:}{end}
    \LinesNumbered
    \Func{\sc Global Align($C, \Delta T$)}{
        \SetKwInOut{Input}{Input}
        \SetKwInOut{Output}{Output}
        \Input{Cost matrix  $C\in\mathbb{R}^{N\times N}$,\\ time offset matrix $\Delta T\in\mathbb{Z}^{N\times N}$}
        \Output{Globally aligning time offsets $\Delta t\in\mathbb{Z}^N$}
        Globally aligning time offsets $\Delta t=\mathbf{0}\in\mathbb{Z}^N$\;
        Globally aligned index group $\mathbf{G} = \varnothing$\;
        Locally aligned index group list $\mathbf{G}_{l}=\varnothing$\;
        Index list $\mathbf{I} = \{(i,j)\vert \forall i<j\}$\;
        $\mathbf{I}\gets \text{\sc Sort}(\mathbf{I})$\Comment{increasing order w.r.t. ${C}_{ij}$}\;
        $(i,j)\gets \mathbf{I}[0]$\Comment{anchor indices}\;
        $\Delta t[i] \gets 0, \Delta t[j] \gets \Delta{T}_{ij}$\;
        Insert $(i,j)$ to $\mathbf{G}$\;
        \For{$k=\{1,\cdots,N(N-1)/2-1\}$}{
            $(i,j)\gets \mathbf{I}[k]$\;
            \uIf{$i\in\mathbf{G}, j\in\mathbf{G}$}{
                \Continue
            }
            \uElseIf{$i\in\mathbf{G}, j\in\mathbf{G}_{{l}}[k],\exists k$}{
                Pop $\mathbf{G}_{{l}}[k]$ and add to $\mathbf{G}$ after shift $\Delta{T}_{ij}$\;
            }
            \uElseIf{$i\in\mathbf{G},j\notin\mathbf{G},j\notin\mathbf{G}_{{l}}[k],\forall k$}{
                Add $j$ to $\mathbf{G}, \Delta{t}[j]\gets \Delta{t}[i] + \Delta{T}_{ij}$\;
            }
            \uElseIf{$i\in\mathbf{G}_{l}[k], j\notin\mathbf{G},j\notin\mathbf{G}[l],\forall l$}{
                Add $j$ to $\mathbf{G}_{{l}}[k]$, $\Delta{t}[j]\gets \Delta t[i] + \Delta{T}_{ij}$\;
            }
            \uElseIf{$i,j\notin\mathbf{G},\notin\mathbf{G}_{l}[k],\forall k$}{
                Add $(i,j)$ to new group in $\mathbf{G}_{{l}}$\;
                $\Delta{t}[i]\gets 0, \Delta{t}[j]\gets \Delta\mathbf{T}_{ij}$\;
            }
            \uElseIf{$i\in\mathbf{G}_{l}[k],j\in\mathbf{G}_{l}[l]$}{
                Pop $\mathbf{G}_{l}[l]$ and add to $\mathbf{G}_{l}[k]$ after shift $\Delta{T}_{ij}$\;
            }
            \uElseIf{$i,j\in\mathbf{G}_{l}[k]$}{
                \Continue
            }
            \Else{
                vice versa for reverse case of $(i,j)$\;
            }
        }
        \KwRet $\Delta t$
    }
    \caption{Global time offset alignment}
    \label{alg:align_time}
\end{algorithm}

We provide detailed pseudocode of the global sequence alignment of whole human motions for time offset estimation in~\cref{alg:align_time}.

\subsection{Procrustes Alignment}
\label{subsec:procrustes}

\DontPrintSemicolon
\SetCommentSty{algcommentfont}
\SetKwFunction{KwFn}{Fn}
\SetKw{Continue}{continue}
\SetNlSty{}{}{}

\begin{algorithm}[t]
    \small
    \SetKwInOut{Input}{Input}
    \SetKwInOut{Output}{Output}
    \SetKwProg{Func}{Function}{:}{end}
    \LinesNumbered
    \Func{\sc Procrustes($X, Y$)}{
        \SetKwInOut{Input}{Input}
        \SetKwInOut{Output}{Output}
        \Input{Point set to align $X=\{\mathbf{x}_i|\mathbf{x}_i\in\mathbb{R}^3\}_{i=1}^N$, \\
        Reference point set $Y=\{\mathbf{y}_i|\mathbf{y}_i\in\mathbb{R}^3\}_{i=1}^N$}
        \Output{scale $s_x, s_y$, translation $\mathbf{t}_x, \mathbf{t}_y$, rotation R}
        \BlankLine
        $\mathbf{t}_x \gets \sum\mathbf{x}_i / N, \mathbf{t}_y \gets \sum\mathbf{y}_i / N$\;
        $s_x \gets \sqrt{\sum\lVert\mathbf{x}_i-\mathbf{t}_x\rVert_2^2/N}$\;
        $s_y \gets \sqrt{\sum\lVert\mathbf{y}_i-\mathbf{t}_y\rVert_2^2/N}$\;
        $\hat{X} \gets \frac{1}{s_x}([\mathbf{x}_i]-\mathbf{t}_x)$\;
        $\hat{Y} \gets \frac{1}{s_y}([\mathbf{y}_i]-\mathbf{t}_y)$\;
        $U, \Sigma, V^\ast \gets \text{SVD}(\hat{Y}\hat{X}^\top)$\;
        $R\gets UV^\ast$\;
        \KwRet $s_x, s_y, \mathbf{t}_x, \mathbf{t}_y, R$
    }
    \caption{Procrustes analysis}
    \label{alg:procrustes}
\end{algorithm}
As we describe in Eq. (7) in the main manuscript, we estimate similarity transform between two 3D joint positions.
We first estimate scale, translation, and rotation that align target joint positions $(\mathbf{J}^i_{\text{global}, t+\Delta t^i};t)$ to the reference joint positions of anchor index $\alpha$, $(\mathbf{J}^\alpha_{\text{global}, t+\Delta t^\alpha};t)$ with Procrustes analysis,
\begin{equation}
\begin{split}
    s_i, s_\alpha, \mathbf{t}_i, \mathbf{t}_\alpha, R = \text{\sc Procrustes}((\mathbf{J}^i_{\text{global}, t+\Delta t^i};t),\\ (\mathbf{J}^\alpha_{\text{global}, t+\Delta t^\alpha};t)).
\end{split}
\end{equation}
We describe details of the Procrustes analysis in~\cref{alg:procrustes}.
Then we can obtain camera poses in the global coordinate (i.e., camera coordinate of anchor index) by applying estimated transformation to the camera poses in the $i$th camera's coordinate, $R^i, \tau^i$ similar to line 3-7 in~\cref{alg:align_cam}.

\subsection{Evaluation Details}
\label{subsec:evaluation_details}

\DontPrintSemicolon
\SetCommentSty{algcommentfont}
\SetKwFunction{KwFn}{Fn}
\SetKw{Continue}{continue}
\SetNlSty{}{}{}

\begin{algorithm}[t]
    \small
    \SetKwInOut{Input}{Input}
    \SetKwInOut{Output}{Output}
    \SetKwProg{Func}{Function}{:}{end}
    \LinesNumbered
    \Func{\textsc{Align Cam}($\{R^i_{\text{est}}, \tau^i_{\text{est}}\}, \{R^i_{\text{ref}}, \tau^i_{\text{ref}}\}$)}{
        \SetKwInOut{Input}{Input}
        \SetKwInOut{Output}{Output}
        \Input{Estimated camera poses $\{R^i_{\text{est}}, \tau^i_{\text{est}}\}$,\\
        Reference camera poses $\{R^i_{\text{ref}}, \tau^i_{\text{ref}}\}$}
        \Output{Aligned estimated camera poses $\{\Tilde{R}^i_{\text{est}}, \Tilde{\tau}^i_{\text{est}}\}$}
        \BlankLine
        Estimated camera centers $\mathbf{o}^i_{\text{est}} \gets -{R^{i\top}_\text{est}}\tau^i$\;
        Reference camera centers $\mathbf{o}^i_{\text{ref}} \gets -{R^{i\top}_\text{ref}}\tau^i$\;
        $s_{\text{est}}, s_{\text{ref}}, \mathbf{t}_{\text{est}}, \mathbf{t}_{\text{ref}}, R \gets \text{\sc Procrustes}(\{\mathbf{o}^i_{\text{est}}\}, \{\mathbf{o}^i_{\text{ref}}\})$\;
        $\Tilde{\mathbf{o}}^i_{\text{est}}\gets s_{\text{ref}}R(\frac{1}{s_{\text{est}}}(\mathbf{o}^i_{\text{est}}-\mathbf{t}_{\text{est}}))+\mathbf{t}^i_{\text{ref}}$\;
        $\Tilde{R}^i_{\text{est}}\gets R^i_{\text{est}}R^\top$\;
        $\Tilde{\tau}^i_{\text{est}}\gets -\Tilde{R}^{i\top}_{\text{est}}\Tilde{\mathbf{o}}_{\text{est}}^i$\;
        \KwRet $\Tilde{R}^i_{\text{est}}, \Tilde{\tau}^i_{\text{est}}$
    }
    \caption{Align cameras}
    \label{alg:align_cam}
\end{algorithm}
In this section, we provide additional details for evaluation of our method.
Since we only optimize camera poses and time offsets of training videos, we do not have accurate poses and time offsets of test videos in the coordinate system that we are optimizing training camera poses and time offsets.
Therefore, we first transform ground-truth test camera poses by aligning the ground-truth training camera poses to the estimated training camera poses.
Starting from the transformed test camera poses, we further optimize camera poses while freezing NeRF parameters with supervision of test view video frames before measuring errors of rendered images.

\label{sec:additional_results}
\begin{figure}
    \centering
    \includegraphics[width=\linewidth]{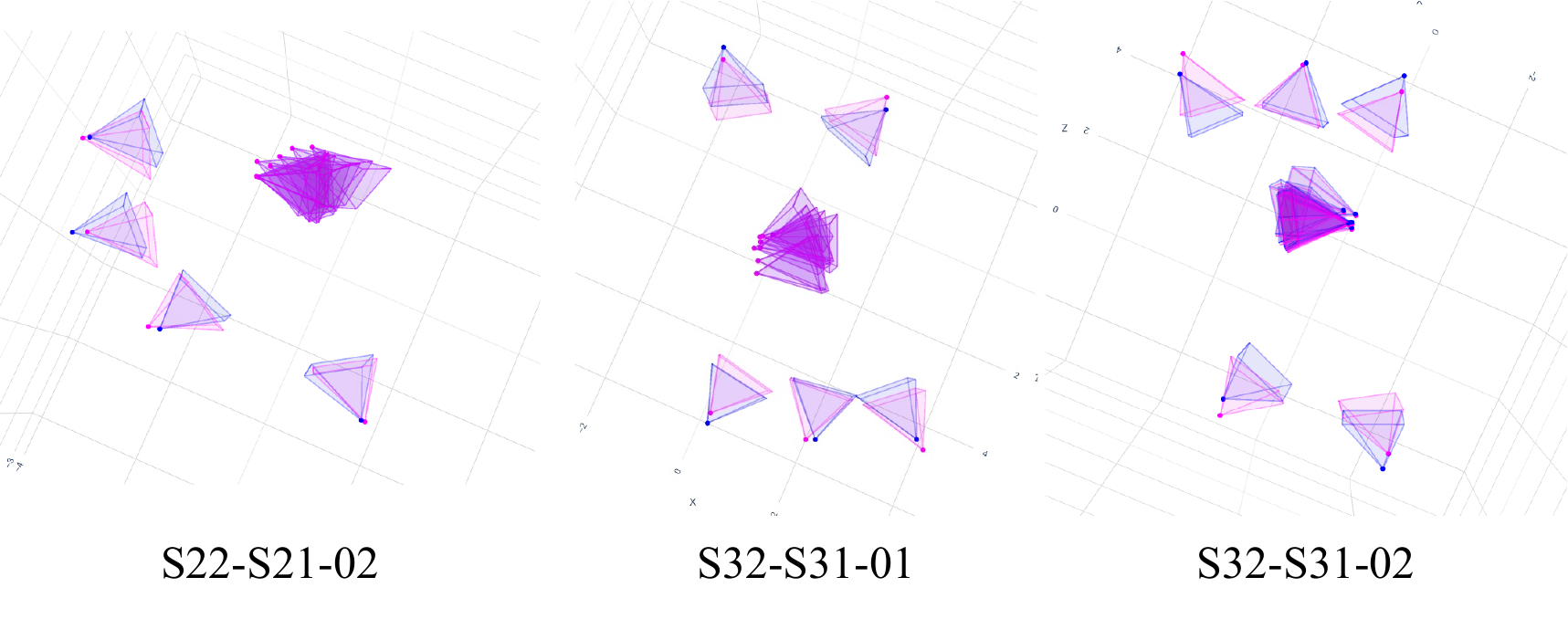}
    \caption{Visual illustration of estimated camera poses from our initialization stage on EgoBody dataset. Red and blue frustums are the ground-truth and estimated camera poses, respectively.}
    \label{fig:pose_egobody}
\end{figure}

\begin{figure}
    \centering
    \includegraphics[width=\linewidth]{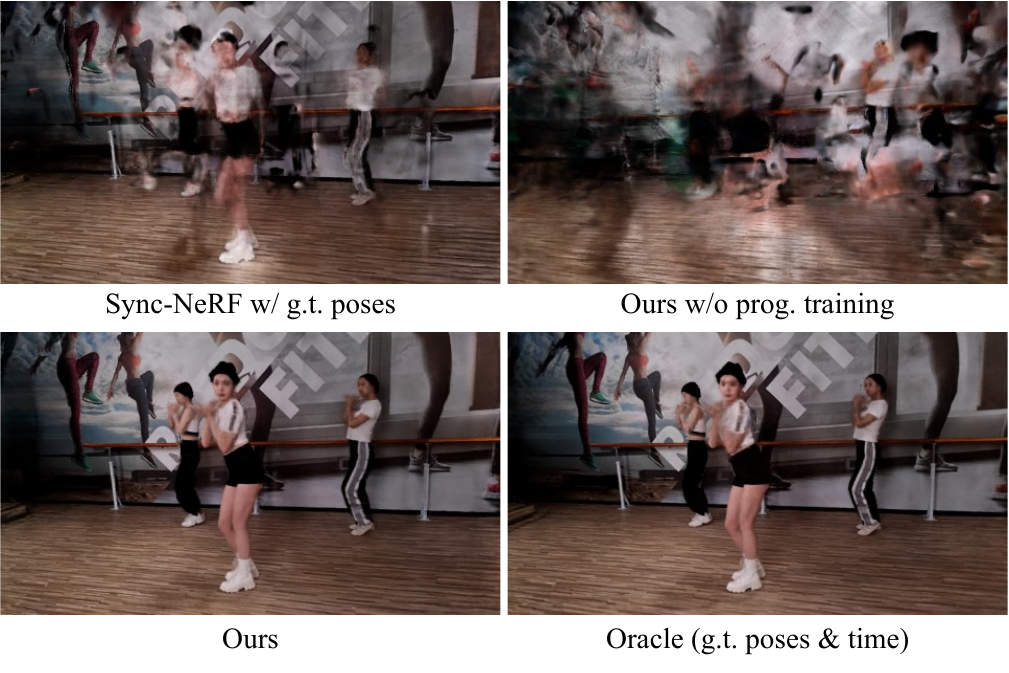}
    \caption{Additional qualitative results on Mobile-Stage dataset.}
    \label{fig:mobile-stage_additional}
\end{figure}
Since the estimated camera poses are up to 3D similarity transformation (scale, rotation, and translation), we align our estimated camera poses to the ground-truth training camera poses before measuring pose errors.
Detailed description of camera alignment procedures used in both novel-view synthesis performance measurement and camera pose accuracy can be found in~\cref{alg:align_cam}.

\section{Additional Results}
\begin{figure*}
    \centering
    \includegraphics[width=\linewidth]{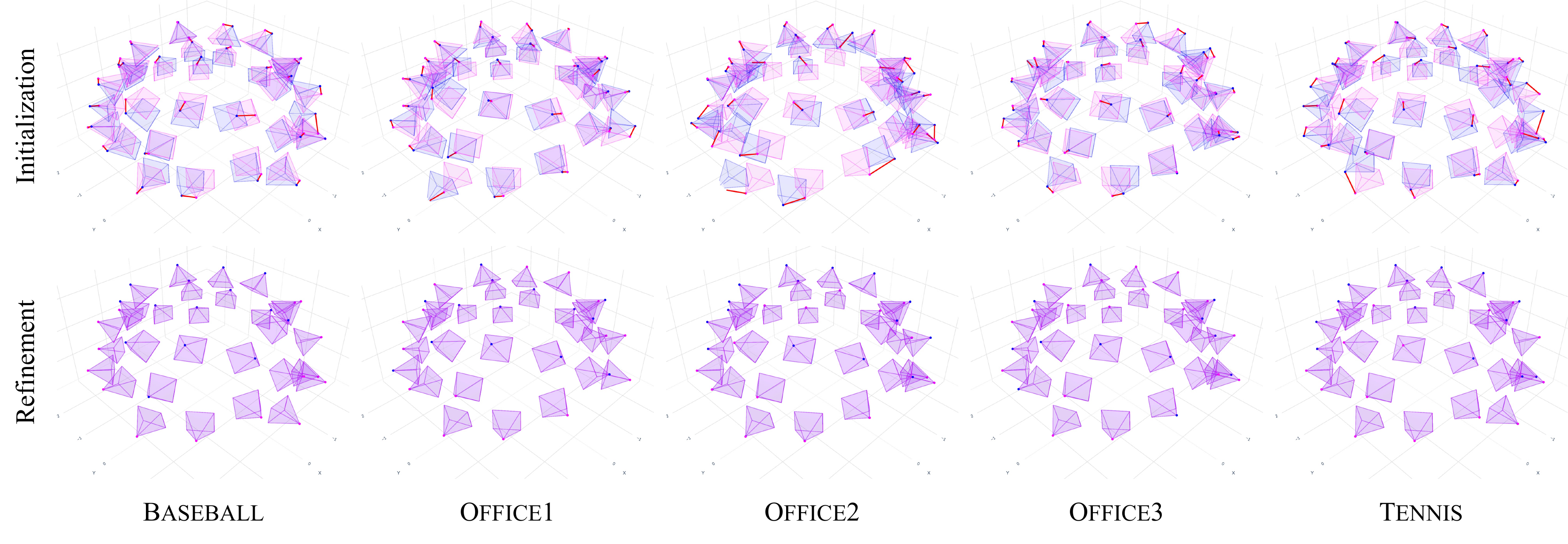}
    \caption{We demonstrate camera pose estimation results of the initialization stage on Panoptic Studio datsaet at the top row (Initialization) and the final results of the joint optimization with K-Planes at the bottom row (Refinement).
    Red frustums are the ground-truth camera poses and blue frustums are the estimated camera poses.}
    \label{fig:pose_refinement_all}
\end{figure*}

We additionally visualize both initialized and refined camera poses in all of the scenes in EgoBody dataset in~\cref{fig:pose_egobody} and Panoptic Studio in~\cref{fig:pose_refinement_all}.
We can observe that our initialization step produces good initial points, and our joint optimization with dynamic NeRF produces near-perfect pose alignments across all scenes.

\begin{figure*}
    \centering
    \includegraphics[width=\linewidth]{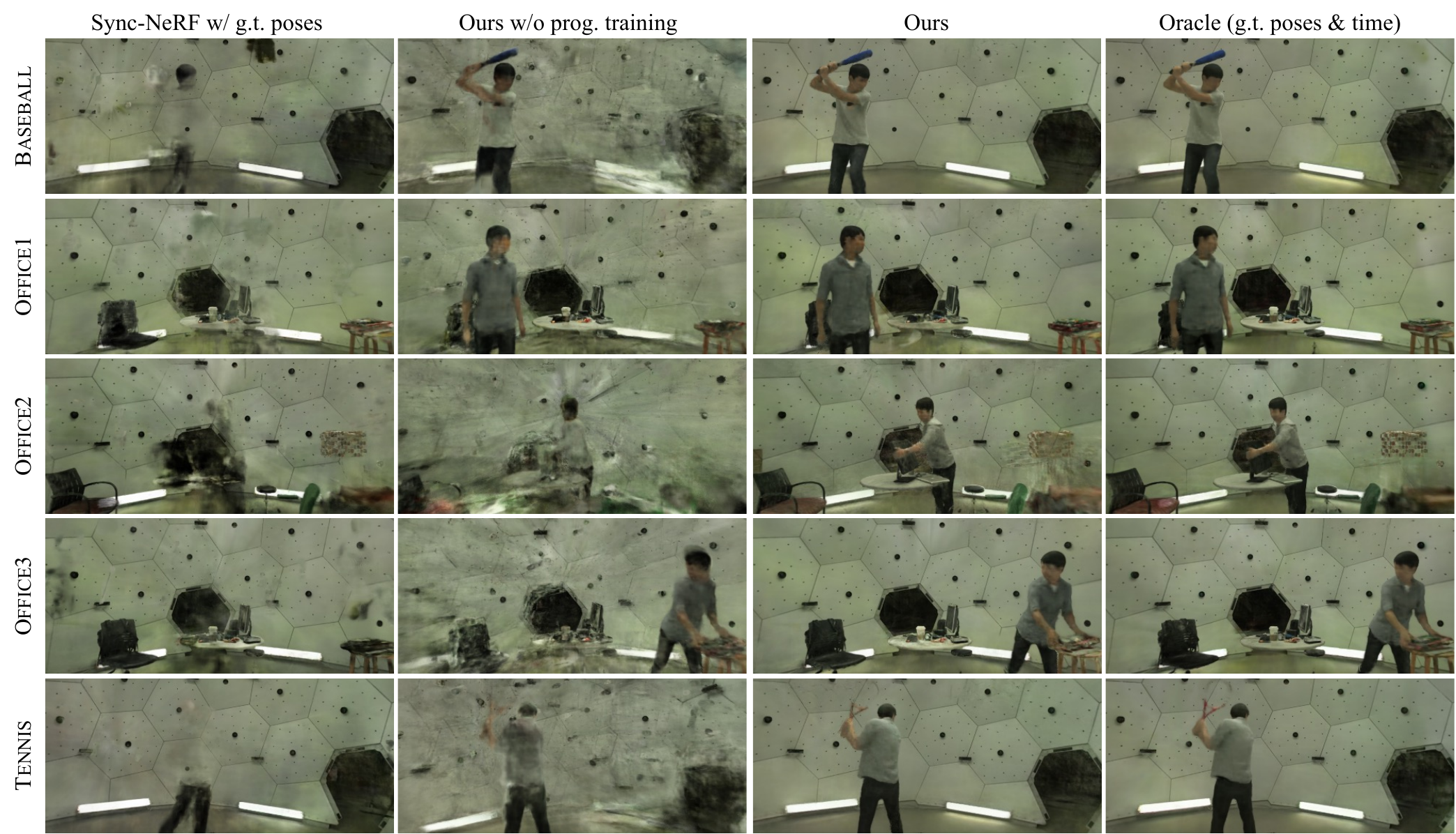}
    \caption{Additional qualitative comparison of novel view synthesis performance.}
    \label{fig:qualitative_additional}
\end{figure*}

Furthermore, we show additional qualitative comparisons in Panoptic Studio dataset in~\cref{fig:qualitative_additional} and Mobile-Stage dataset in~\cref{fig:mobile-stage_additional}.
We also provide videos rendered at the test viewpoint in the supplementary material.
We recommend the readers to see the videos.


\end{document}